%% file: main.tex
\documentclass[10pt,twocolumn,letterpaper]{article}

\usepackage{cvpr}              

\input{preamble}
\usepackage[accsupp]{axessibility}  
\definecolor{cvprblue}{rgb}{0.21,0.49,0.74}
\usepackage[pagebackref,breaklinks,colorlinks,allcolors=cvprblue]{hyperref}
\usepackage{pifont}
\usepackage{multirow}

\def\paperID{5002} 
\def\confName{CVPR}
\def\confYear{2025}

\newcommand{\methodAbbr}{SnapGen-V}

\title{
\methodAbbr: 
Generating a Five-Second Video within Five Seconds on a Mobile Device
}

\author{
Yushu Wu\affmark[1,2\textdagger]\quad
Zhixing Zhang\affmark[1,3\textdagger]\quad
Yanyu Li\affmark[1\textdagger\textdaggerdbl]\quad
Yanwu Xu\affmark[1]\quad
Anil Kag\affmark[1]\quad 
Yang Sui\affmark[1]\quad \\
Huseyin Coskun\affmark[1]\quad
Ke Ma\affmark[1]\quad
Aleksei Lebedev\affmark[1]\quad
Ju Hu\affmark[1]\quad 
Dimitris N. Metaxas\affmark[3]\quad \\
Yanzhi Wang\affmark[2]\quad
Sergey Tulyakov\affmark[1]\quad
Jian Ren\affmark[1\textdaggerdbl] \\
\affmark[1]Snap Inc.\quad\quad \affmark[2]Northeastern University \quad\quad \affmark[3]Rutgers University \\
}

\begin{document}
\twocolumn[{
\renewcommand\twocolumn[1][]{#1}
\maketitle
\input{figs/0_teaser}
}]

\begingroup
\def\thefootnote{\textdagger}\footnotetext{Equal contribution}
\def\thefootnote{\textdaggerdbl}\footnotetext{Corresponding authors}
\endgroup

\input{sec/0_abstract}  

\input{sec/1_intro}

\input{sec/2_related}
\input{sec/3_method}
\input{sec/4_experiments}
\input{sec/5_conclusion}

{
    \small
    \bibliographystyle{ieeenat_fullname}
    \bibliography{main}
}

\input{sec/X_suppl}

\end{document}

%% file: preamble.tex
\usepackage{color, colortbl}
\definecolor{LightCyan}{rgb}{0.88,1,1}

\usepackage{algorithm}
\usepackage{algpseudocode}
\newcommand*{\affmark}[1][*]{\textsuperscript{#1}}

%% file: figs/0_teaser.tex


\begin{center}
    \centering
    \captionsetup{type=figure}
    \includegraphics[width=1.0\linewidth]{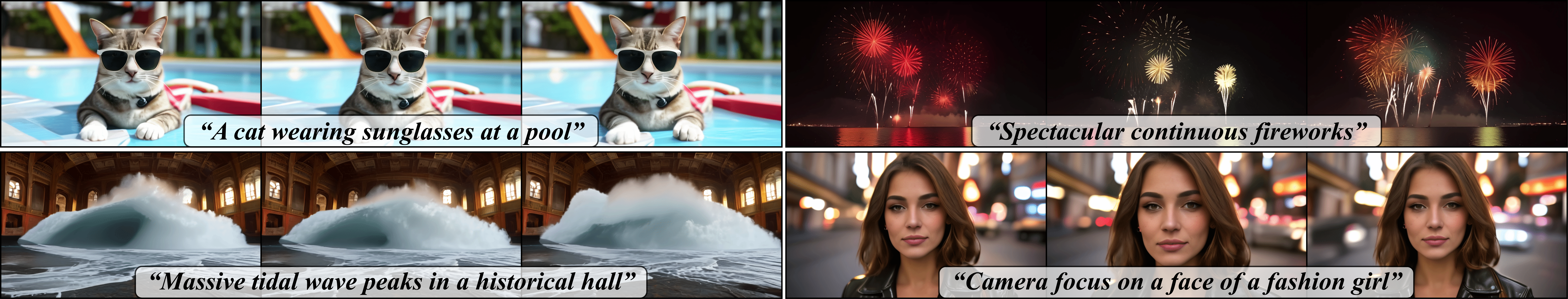}
    \caption{\textbf{Example generation results from our mobile text-to-video model.} Our model can generate high-quality and motion consistent $5$-second videos on a \emph{mobile} device (\eg, iPhone 16 Pro Max) within $5$ seconds.}
    \label{fig: teasor}
\end{center}

%% file: sec/0_abstract.tex
\begin{abstract}

We have witnessed the unprecedented success of diffusion-based video generation over the past year. 
Recently proposed models from the community have wielded the power to generate cinematic and high-resolution videos with smooth motions from arbitrary input prompts. 
However, as a supertask of image generation, video generation models require more computation and are thus hosted mostly on cloud servers, limiting broader adoption among content creators. 
In this work, we propose a comprehensive acceleration framework to bring the power of the large-scale video diffusion model to the hands of edge users. 
From the network architecture scope, we initialize from a compact image backbone and search out the design and arrangement of temporal layers to maximize hardware efficiency. 
In addition, we propose a dedicated adversarial fine-tuning algorithm for our efficient model and reduce the denoising steps to $4$. 
Our model, with only $0.6$B parameters, can generate a $5$-second video on an iPhone 16 PM within $5$ seconds.
Compared to server-side models that take minutes on powerful GPUs to generate a single video, we accelerate the generation by magnitudes while delivering on-par quality.
Project page at~\href{https://snap-research.github.io/snapgen-v/}{https://snap-research.github.io/snapgen-v/}.

\end{abstract}

%% file: sec/1_intro.tex
\section{Introduction}
\label{sec:intro}

Recently, the rapid advancement of video diffusion models \cite{blattmann2023align} inspires revolutions in content creation. 
With the emergence of video models from industry \cite{polyak2024movie} and research community \cite{yang2024cogvideox}, content creators can animate a static image \cite{blattmann2023stable} or generate cinematic videos from arbitrary prompts \cite{yang2024cogvideox,genmo2024mochi,zhou2024allegro}. 
Video diffusion models also enable downstream applications like video editing \cite{jeong2024dreammotion,feng2025wave,liang2024flowvid}, novel view synthesis \cite{voleti2025sv3d,kwak2024vivid}, and multi-modal generation \cite{team2023gemini}. 

Despite the success in generation quality, the huge number of parameters and slow generation speed prohibit the wide deployment of video diffusion models. 
For instance, CogVideoX-5B~\cite{yang2024cogvideox} generates a video~($49$ frames at $8$ fps, $720\times480$~resolution) in $5$ minutes with $50$ inference steps on an NVIDIA A100 GPU. 
Compared to text-to-image diffusion models \cite{rombach2022high}, video diffusion models require extra parameters to model sophisticated motions \cite{wu2023tune,guo2023animatediff,blattmann2023stable}. 
In addition, video data usually incurs higher spatial-temporal resolution for UNet denoisers \cite{guo2023animatediff,blattmann2023stable}, or equivalently more tokens for DiT models \cite{peebles2023scalable}, which adds up to the computation complexity. 
Recent works explore efficient model architectures and attention mechanisms for image diffusion models~\cite{yuan2024ditfastattn,chen2024edt, xie2024sana}. 
However, there is little effort in the literature dedicated to accelerating and deploying video models at scale, especially for mobile devices.

In this work, we systematically investigate the redundancies in video diffusion models and propose a mobile acceleration framework. 
First, we obtain an efficient spatial backbone by following prior works~\cite{li2024snapfusion,fang2023structural} to prune a pre-trained text-to-image diffusion model \cite{rombach2022high}. 
The pruned model achieves $2.5\times$ size compression and more than $10 \times$ speedup compared to Stable Diffusion v1.5 \cite{rombach2022high}, while maintaining comparable generative quality. 
Starting from a pre-trained image model offers two key benefits: (i) it eliminates the need for costly large-scale pre-training, and (ii) with a compact image model, we can significantly narrow the search space in subsequent stages, focusing only on optimizing temporal layers, and thereby accelerating the discovery of the final model architecture.

Even with the efficient image backbone, applying previous temporal inflation methods \cite{blattmann2023align,guo2023animatediff,blattmann2023stable} still results in tremendous computation cost and encounters out-of-memory issues on mobile. 
Thus, our second stage is to systematically investigate different types of temporal layers and perform a latency-memory joint search to determine the spatial-temporal architecture for efficient mobile deployment. 
Unlike previous methods \cite{blattmann2023align,guo2023animatediff,blattmann2023stable} that typically rely on a specific type of temporal modeling layer, we investigate all possible designs, including temporal attention, spatial-temporal full attention (3D attention), temporal cross attention, and temporal convolutions (Conv3D). 
Besides, our search space includes the position (resolution) to apply these temporal layers, and number of layers to use (in \cref{sec:method:model}). 
We profile the computation, memory footprint, and on-device latency of architecture candidates, and perform evolutionary search to discover the architecture with Pareto optimality of speed and quality.
The searched network is only $0.6$B in size, and can generate a $5$-second video clip on iPhone 16 PM without hitting memory bound. 
In contrast, all prior video diffusion models fail to run on mobile, even for the smallest open-sourced ones like the $16$-frame AnimateDiff \cite{guo2023animatediff} and $14$-frame SVD \cite{blattmann2023stable} (\cref{table: parameter latency comparison}).

Finally, to further speed up generation on mobile, we distill our efficient video diffusion model with a tailored adversarial fine-tuning method capable of image-video mixed training. 
We reduce the number of denoising steps from $25$ to $4$, and eliminate classifier-free guidance~\cite{ho2022classifier}, leading to more than $12\times$ speedup without performance degradation. 
As in \cref{table: parameter latency comparison}, our \emph{mobile} speed is faster than most GPU-deployed (\eg, on A100) counterparts. 

\input{tables/parameter_latency_comparison}

With the proposed framework, we successfully deploy our $0.6$B text-to-video model on an iPhone 16 Pro Max, achieving the generation of a $5$-second video clip within $5$ seconds. 
This work represents not only the very \emph{first} mobile deployment attempt of the video diffusion model, but also demonstrates its real-time potential\footnote{This work performs conventional T2V generation, generating an entire video at once. However, a more suitable real-time capability would involve streamlined, continuous video generation, which we leave for future work. }. 
Our contributions are summarized as follows:
\begin{itemize}
    \item Through image-video joint training, spatial and temporal architecture design, and mobile-driven latency-memory joint architecture search, we develop a comprehensive mobile acceleration framework for the text-to-video diffusion model.
    \item We propose an adversarial fine-tuning technique tailored for video diffusion models. Despite the already compact nature of our mobile denoiser, we further distill it to $4$ denoising steps with superior quality. 
    \item Our work is the very first one to show the possibility of real-time text-to-video generation on mobile devices, unlocking the possibility of deploying applications of video diffusion models at scale. 
\end{itemize}

%% file: tables/parameter_latency_comparison.tex
\begin{table}[ht]
\centering
\resizebox{\columnwidth}{!}{%
\renewcommand\tabcolsep{2pt}
\small
\begin{tabular}{lcccccc}
\toprule
Model     & Type & Steps & Params~(B) & A100~(s) & iPhone~(s) & Vbench ($\uparrow$)\\
\midrule
OpenSora-v1.2     & DiT & 30 & 1.2     & 31.00   & \ding{55}  &    79.76 \\
CogVideoX-2B      & DiT & 50 & 1.6     & 54.09   & \ding{55}  &   80.91  \\
AnimateDiff-V2    & UNet & 25 & 1.2     & 9.04    & \ding{55}  &   80.27  \\
AnimateDiffLCM    & UNet & 4 & 1.2     &  1.77   & \ding{55}  &   79.42  \\
\hline
Ours             & UNet & 4 & 0.6     & 0.47    &  4.12      &    81.14   \\
\bottomrule
\end{tabular}
}
\caption{Comparison of size (number of parameters), speed (tested on NVIDIA A100 and iPhone 16 Pro Max), and performance (on VBench~\cite{huang2023vbench}) for various models. }
\label{table: parameter latency comparison}
\end{table}


%% file: sec/2_related.tex
\section{Related Work}
\label{sec:related}
\noindent\textbf{Video Diffusion Models.}
Denoising Diffusion Probabilistic Models \cite{ho2020denoising} is the trending paradigm for building video diffusion models, demonstrating photorealistic quality and generic generation capabilities. 
Pioneer works often start from a pre-trained text-to-image diffusion model \cite{rombach2022high} and insert temporal layers to model motions along frame sequence \cite{blattmann2023align,wu2023tune,guo2023animatediff,blattmann2023stable, zhang2024avid}. 
In addition, training-free noise tuning techniques are proposed to ease the alignment between frames \cite{wu2025freeinit,qiu2023freenoise,kim2024fifo,lu2024freelong,zhanfast}. 
Later, with the emergence of large-scale, high-quality video datasets \cite{chen2024panda,nan2024openvid} and Transformer backbones \cite{peebles2023scalable}, subsequent works curate their own dataset and build large video diffusion models with exceptional quality, such as the open-sourced CogVideoX~\cite{yang2024cogvideox}, Mochi 1~\cite{genmo2024mochi}, PyramidalFlow \cite{jin2024pyramidalflow}, Allegro \cite{zhou2024allegro}, and close-sourced ones including Hailuo \cite{hailuo}, Runway Gen 3 Alpha \cite{gen3a}, Kling \cite{kling}, Luma Dream Machine \cite{luma}, Pika 1.5 \cite{pika1.5}, Sora \cite{sora}, and MovieGen~\cite{polyak2024movie}. 
Remarkably, the open-sourced projects Open-Sora \cite{opensora} and Open-Sora-Plan \cite{pku_yuan_lab_and_tuzhan_ai_etc_2024_10948109} provide the community with reliable implementations to replicate large-scale video diffusion models. 

Dividing by task type, video diffusion models can be categorized into text-to-video generation \cite{guo2023animatediff,hong2022cogvideo,yang2024cogvideox,genmo2024mochi,li2024t2vturbo,li2024t2vturbov2,jin2024pyramidalflow,zhou2024allegro,team2023gemini,chen2023videocrafter1,chen2024videocrafter2,qing2024hierarchical,gupta2023photorealistic}, image-to-video generation \cite{blattmann2023stable,zhou2024storydiffusion}, or specific motion controls \cite{zeng2024make,guo2025sparsectrl,ren2024customize,zhao2025motiondirector,wu2024motionbooth}.
Though some work \cite{li2024snapfusion,yuan2024ditfastattn,chen2024edt,zhanfast} aim to improve the efficiency of diffusion models, the acceleration for mobile deployment of video diffusion models is still in absent. Popular video models \cite{yang2024cogvideox,jin2024pyramidalflow,genmo2024mochi} can only run on server-level GPUs to generate videos in tens of seconds or even minutes. 







\noindent\textbf{Step Distillation} brings almost linear generation speedup for diffusion models. 
Early work \cite{salimans2022progressive,li2024snapfusion} progressively distill a student network to predict a further ODE location with teacher guidance, resulting in fewer inference steps, while Consistency Models \cite{song2023consistency,song2023improved} and Rectified Flow \cite{liu2022flow} refine the prediction objective to clean data or global velocity to achieve fewer-step inference. 
Later works \cite{ufogen,add,ladd} further incorporate adversarial loss to distill a single-step student, and enhance multi-step results as well. 

Despite extensive research in image diffusion models \cite{dmd,dmd2,cfm,wang2024rectified,kim2024simple,mei2024codi,dao2025swiftbrush}, step distillation for video diffusion model is under-explored. One type of work applies consistency distillation to generate videos in 4 steps~\cite{videolcm,wang2024animatelcm,zhai2024motion}. Another trend adopts adversarial distillation to achieve few-step (1-2) generation~\cite{lin2024animatediff,sfv,mao2024osv}.
However, these methods distill pre-trained large models with enough redundancy along the trajectory, while we find them not applicable to our efficient model and yield inferior performance. 


%% file: sec/3_method.tex
\section{Method}
\label{sec:method}
\input{figs/1_overview}


Our objective is to achieve high-fidelity and temporally consistent video generation 
on mobile devices.
However, current text-to-video diffusion models face two key challenges in reaching this goal: (a) the memory and computation requirement is beyond the capability of even the most powerful mobile chips, \ie iPhone A18 Pro, and 
(b) denoising with dozens of steps to generate a single output further slows down the process. 
To address these challenges, we propose a three-stage framework to accelerate video diffusion models on the mobile platform.
First, we prune from a pre-trained text-to-image diffusion model to obtain an efficient spatial backbone. 
Second, we inflate the spatial backbone with a novel combination of temporal modules which are searched out with our mobile-oriented metrics. 
Finally, through adversarial training, our efficient model attains the capability to generate high-quality videos in only $4$ steps.

\subsection{Preliminaries}
\label{sec:method:preliminaries}

Following ~\cite{opensora}, we employ a spatial-temporal VAE to compress image and video data into the latent space. 
Given video or image data $\mathbf{v} \in \mathbb{R}^{n \times 3 \times H \times W}$, where $n$ is the number of frames with height $H$ and width $W$, the spatial-temporal encoder, $\mathbf{E}$, maps the data to a latent space. 
The encoded frames are represented as $\mathbf{x}_0 = \mathbf{E}(\mathbf{v})$, resulting in $\mathbf{x}_0 \in \mathbb{R}^{\tilde{n} \times 4 \times \tilde{H} \times \tilde{W}}$. 
Here, $\mathbf{x}_0 \sim p_{data}(\mathbf{x}_0)$ is a $4$-channel latent,
with a temporal compression of $\tilde{n}=n/4$, and spatial compression as $\tilde{H}=H/8$, $\tilde{W}=W/8$.

We follow Rectified Flow ~\cite{wang2024rectified} to train our latent diffusion model. 
According to the flow-matching-based diffusion form, the intermediate noisy state $\mathbf{x}_t$ is defined as:
\begin{equation}
    \label{equ:forward}
    \mathbf{x}_t = \left(1 - t\right)\mathbf{x}_0 + t \epsilon, \text{where}~\epsilon \sim \mathcal{N}\left(0, \mathit{I}\right),
\end{equation}
which is a linear interpolation between the data distribution and a standard normal distribution.
The model aims to learn a vector field $v_\theta\left(t, \mathbf{x}_t\right)$ using the Conditional Flow Matching objective, \ie,
\begin{equation}
    \label{equ:loss:fm}
    \mathcal{L} = \mathbb{E}_{t, \epsilon, \mathbf{x}_0}\left\Vert v_\theta\left(t, \mathbf{x}_t\right) - u_t\left(\mathbf{x}_t \vert \mathbf{x}_0\right) \right\Vert_2^2,
\end{equation}
where $u_t\left(\mathbf{x}_t \vert \mathbf{x}_0\right) = \epsilon - \mathbf{x}_0$. 
Following~\cite{sd3}, during training, we sample $t$ from a logit-normal distribution, \ie,
\begin{equation}
    \label{equ:logitnorm}
    \pi\left(t; m, s\right) = \frac{1}{s\sqrt{2\pi}}\frac{1}{t\left(1 - t\right)}\operatorname{exp}\left(-\frac{\left(\operatorname{logit}\left(t\right) - m\right)^2}{2s^2}\right),
\end{equation}
where $\operatorname{logit}(t) = \operatorname{log}\frac{t}{1 - t}$, $m$ and $s$ are the location parameter and scale parameter, respectively.

\subsection{Hardware Efficient Model Design}
\label{sec:method:model}

\noindent \textbf{Spatial Backbone.}
\label{sec:spatial design}
We follow ~\cite{li2024snapfusion,fang2023structural} to first prune an efficient text-to-image model as the spatial backbone. 
Specifically, we start from Stable Diffusion v1.5 ~\cite{rombach2022high}, and borrow the knowledge from prior arts \cite{li2024snapfusion} to remove the most mobile-unfriendly attentions. We then prune the network depth and width following \cite{fang2023structural} and achieve $\times2.5$ size reduction and more than $10 \times$ speedup on mobile devices.
We include qualitative visualizations of our image model in the \emph{supplementary material}. 
Note that we use a UNet denoiser ~\cite{ho2020denoising}, leaving the exploration of DiT ~\cite{peebles2023scalable} to future work. 
The hierarchical structure of the UNet denoiser forms a good search space to achieve mobile efficiency, while the computation complexity of DiT grows quadratically with the number of tokens (generation resolution), making it challenging for mobile deployment. 
\input{figs/2_module_complexity}

\noindent\textbf{Temporal Layer Design.}
Current latent video diffusion models typically adopt temporal self-attentions~\cite{guo2023animatediff}, cross-attentions~\cite{opensora}, and convolutions~\cite{blattmann2023stable} to model temporal dependencies. CogVideoX~\cite{yang2024cogvideox} demonstrates significant performance gain by using full 3D-Attention, at the cost of more computations and memory consumption. 
In this work, we enumerate and investigate all types of temporal modeling methods, including \emph{Conv1D}, \emph{Conv3D}, \emph{SelfAttention1D}, \emph{SelfAttention3D}, 
\emph{CrossAttention1D}, and \emph{CrossAttention3D}, and profile their complexity in~\cref{fig: complexity}. 
For instance, \emph{SelfAttention1D} only models temporal dependency on a single coordinate, while \emph{SelfAttention3D} models global dependencies and has the potential to deliver much stronger performance. 
However, the computation complexity of \emph{SelfAttention3D} grows quadratically with respect to $\tilde{t} \times \tilde{H} \times \tilde{W}$, while \emph{SelfAttention1D} is linear with respect to $\tilde{H}$ and $\tilde{W}$, which makes \emph{SelfAttention3D} much more costly at higher resolutions. 
On the other hand, the computation of \emph{CrossAttentionND} is determined by both spatial-temporal resolution and the number of tokens from the text encoder. 
\emph{Conv1D} and \emph{3D} are locality alternatives for \emph{SelfAttention1D} and \emph{3D}, respectively. 
Though the computation complexity and memory footprint for each design candidate can be easily profiled, as in \cref{fig: complexity}, it is still crucial and challenging to build a spatial-temporal network with optimized arrangements of these operators. 
We propose to perform a latency-memory joint architecture search to determine \emph{which}, \emph{where}, and the \emph{number} of temporal layers to use for our efficient video diffusion model on mobile, as follows.

\noindent\textbf{Latency and Memory Guided Architecture Search.}
\label{sec: method search}
Prior to searching the architecture, we construct a look-up table containing the inference latency and the memory footprint of different temporal layers.
For each candidate 1D or 3D operator~(\ie, \textit{ConvND}, \textit{SelfAttentionND}, \textit{CrossAttentionND}), we benchmark the latency and memory consumption under different spatial-temporal resolutions on hardware. 
We then clean the search space by eliminating OOM states. 
Then we perform evolutionary search to obtain the temporal design with Pareto optimality. 
The architecture candidate is trained on precomputed video latents for $20K$ iterations with the spatial backbone frozen, and is evaluated on VBench \cite{huang2023vbench} to obtain the scores as the quality metric.
We include the detailed search algorithm, action space, and total search time in the \emph{supplementary material}. 

\noindent\textbf{Image-Video Joint Training. }
\label{sec: image video joint training}
Upon the finalized model architecture, we perform image-video joint training under various clip lengths and aspect ratios with all parameters updated for another $100K$ iterations. 
After the joint training, our efficient model is capable of generating videos with various lengths and aspect ratios under a conventional recipe, \ie, $25$ steps with classifier-free guidance. 

\noindent\textbf{VAE Decoder Compression.}
We use the spatial temporal-decoupled VAE from OpenSora \cite{opensora}, which has $4$ latent channels, $8\times 8$ spatial compression and $4\times$ temporal compression. 
To decode a $17$-frame video clip on mobile, the original \cite{opensora} temporal decoder takes 23,100 ms, and the spatial decoder takes 4100 ms, which we found to be a bottleneck for the generation speed. 
To increase the speed, we focus on the decoder only. We freeze the VAE encoders and prune the temporal and spatial decoder on our video and image dataset, respectively. 
Our efficient temporal decoder runs at $210$ ms and spatial decoder at $330$ ms to decode a $17$-frame video clip, reaching $50\times$ speedup with negligible loss in quality. 
Further details about VAE compression are included in the \emph{supplementary material}.

\subsection{Latent Adversarial Fine-tuning}
\label{sec:method:adversarial}

Our training procedure involves two networks: a generator $\mathcal{G}_\theta$ and a discriminator $\mathcal{D}\phi$. 
Similar to prior work~\cite{ladd, sfv}, we initialize our generator with pre-trained diffusion model weights $\theta$, while the discriminator is also partially initialized from $\theta$.
Specifically, the backbone of the discriminator adopts the same architecture and weights as the pre-trained UNet encoder, with these backbone parameters remaining frozen during training. 
Additionally, we enhance the discriminator with spatial-temporal discriminator heads added after each backbone block, with only these head parameters being updated in the discriminator training phase.
As illustrated on the right in \cref{fig: overview}, each discriminator head consists of a spatial ResBlock and a temporal self-attention block. 
This design allows our discriminator to effectively handle both image and video data during fine-tuning. 
We analyze the impact of joint image-video fine-tuning in \cref{sec:exp:ablation}.

For a real data sample $\mathbf{x}_0$, a noisy data point $\mathbf{x}_t$ is generated through a forward diffusion process, as described in \cref{equ:forward}. 
We set intermediate timesteps as $0 < T_k < \cdots < T_1 = 1.0$ and sample $t$ from these timesteps, where $k$ is the number of timesteps selected for generator training (set to $k=4$ in practice).
The generator, then, predicts the velocity at $\mathbf{x}_t$ as $\mathcal{G}_\theta\left(\mathbf{x}_t, t\right)$.

To train the discriminator, we first sample a target timestep $t^\prime$ from a logit-normal distribution, as shown in \cref{equ:logitnorm}.
Using the forward process in \cref{equ:forward}, we obtain the real sample $\mathbf{x}_{t^\prime} = \left(1 - t^\prime\right)\mathbf{x}_0 + t^\prime \epsilon$. 
The fake sample, $\hat{\mathbf{x}}_{t^\prime}$, is computed as $\hat{\mathbf{x}}_{t^\prime} = \mathbf{x}_t + \left(t^\prime - t\right) \cdot \mathcal{G}_\theta\left(t, \mathbf{x}_t\right)$, as shown in \cref{fig: adversarial}.
Following established approaches~\cite{add, sauer2023stylegan, sauer2021projected, sfv}, we employ hinge loss~\cite{lim2017geometric} as the adversarial objective to enhance performance. 
The discriminator's goal is to differentiate between real and fake samples by minimizing:
\begin{equation}
    \label{equ: discriminator adv}
    \begin{split}
    \mathcal{L}_\text{adv}^\mathcal{D} = & 
    \mathbb{E}_{t^\prime, x_0} \left[\max\left(0, 1 + \mathcal{D}_\phi\left(\mathbf{x}_{t^\prime}, t^\prime\right)\right) \right] \\
     + & \mathbb{E}_{t, t^\prime, x_0}\left[\max\left(0, 1 - \mathcal{D}_\phi\left(\hat{\mathbf{x}}_{t^\prime}, t^\prime\right)\right)\right],
    \end{split}
\end{equation}

The adversarial objective for the generator is defined as:
\begin{equation}
    \label{equ: generator adv}
    \mathcal{L}_\text{adv}^\mathcal{G} = 
    \mathbb{E}_{t, t^\prime, x_0}[ \mathcal{D}_\phi\left(\hat{\mathbf{x}}_{t^\prime}, t^\prime\right)].
\end{equation}
Following ~\cite{sfv}, we also incorporate a reconstruction objective to enhance stability, defined as:
\begin{equation}
    \label{equ: reconstruction}
    \mathcal{L}_{\text{recon}} = \sqrt{\left\Vert \hat{\mathbf{x}}_{0} - \mathbf{x}_0 \right\Vert_2^2 + c^2} - c,
\end{equation}
where $\hat{\mathbf{x}}_{0} = \mathbf{x}_t  - t \cdot \mathcal{G}_\theta\left(t, \mathbf{x}_t\right)$, and $c>0$ is an adjustable constant.

\input{figs/2_adversarial}

\noindent \textbf{Discussion.} 
Our latent adversarial training pipeline is inspired by SF-V~\cite{sfv}.
Similar to SF-V, we set $k=4$ and utilize the part of the pre-trained diffusion model as the backbone for the discriminator. 
However, our approach introduces several key differences. 
\emph{First}, our method is built on an efficient UNet specifically designed for mobile devices, with fewer parameters than SVD~\cite{blattmann2023stable}, making it a more challenging task.  
\emph{Second}, we redesign the discriminator heads: instead of using separate spatial and temporal heads, we integrate them into a unified spatial-temporal head for adversarial training. 
Rather than handling the temporal dimension separately with $1$-D convolutional kernels, we incorporate a temporal self-attention layer into the spatial discriminator head after the $2$-D ResBlock, forming a spatial-temporal discriminator head. 
This unified design enables our model to be jointly trained on both image and video data, which, as demonstrated in \cref{sec:exp:ablation}, significantly enhances the performance of the fine-tuned model.

%% file: figs/1_overview.tex
\begin{figure*}
  \centering
  \includegraphics[width=\linewidth]{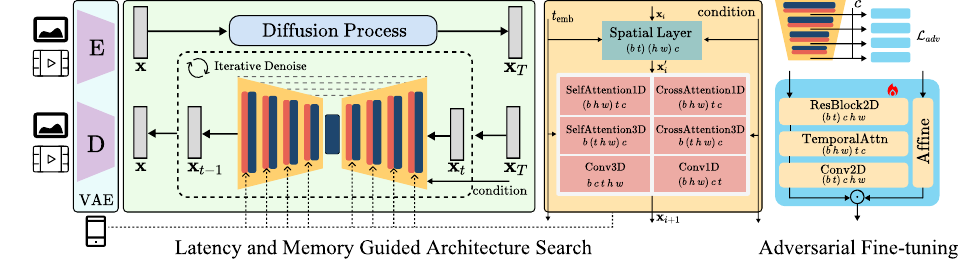}
  \caption{\textbf{Framework Overview.}
  In the Latency and Memory Guided Architecture Search, we freeze the pretrained efficient spatial layers and conduct evolutionary search over temporal layer based on the memory and latency constraint.
  During the Adversarial Fine-tuning stage, we initialize the discriminator with the weights from the text-to-video model trained in the first stage. The discriminator employs the encoder of the UNet as its backbone, which remains \emph{frozen}. We add spatial-temporal discriminator heads after each downsampling block, updating only these heads during training. Following prior works~\cite{sauer2021projected, sauer2023stylegan, sfv}, each head conditions on pooled text embeddings $\mathbf{c}$ projected via a linear layer. Input features are first reshaped to merge the temporal and batch axes for processing through a 2D ResBlock, and then reshaped again to merge spatial dimensions before the temporal self-attention block.}
  \label{fig: overview}
\end{figure*}

%% file: figs/2_module_complexity.tex
\begin{figure}[htb]
  \centering
  \includegraphics[width=\linewidth]{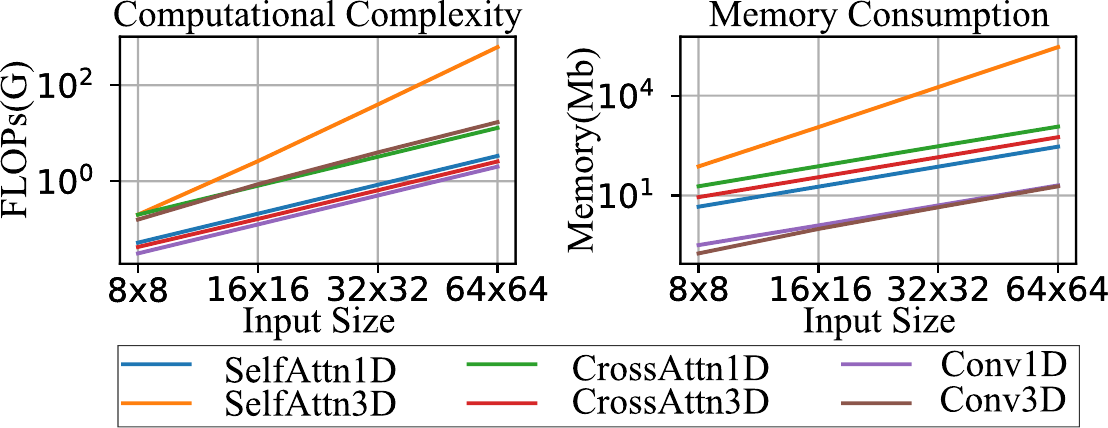}
  \caption{\textbf{Computation Complexity and Memory Consumption Analysis.} The computation complexity and memory consumption of different temporal layer for various input size. The temporal dimension is fixed to $12$ for simplicity.}
  \label{fig: complexity}

  \vspace{-1.em}
\end{figure}

%% file: figs/2_adversarial.tex
\begin{figure}[htb]
  \centering
  \includegraphics[width=0.8\linewidth]{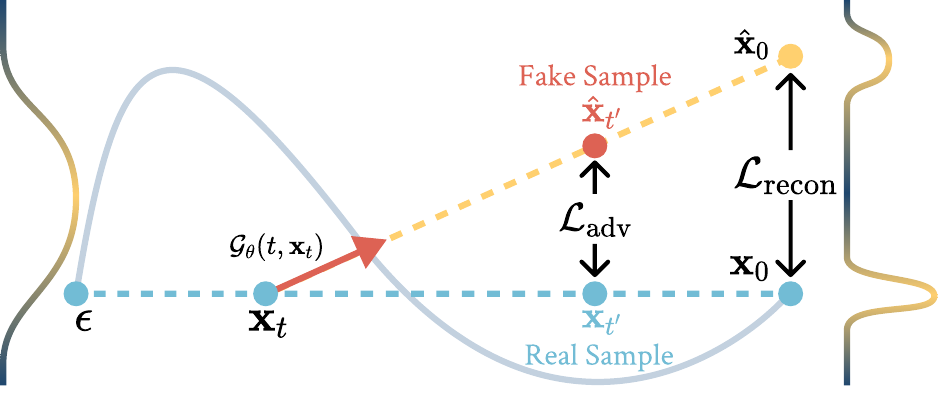}
  \caption{\textbf{Latent Adversarial Fine-tuning.} Given a latent $\mathbf{x}_0$ and a noise latent $\epsilon$, we obtain the intermediate noisy latent $\mathbf{x}_t$ through a forward diffusion process. The generator then predicts the velocity as $\mathcal{G}_\theta\left(\mathbf{x}_t, t\right)$. Using the predicted velocity, we compute $\hat{\mathbf{x}}_0$ and calculate the reconstruction loss $\mathcal{L}_{\text{recon}}$ between $\mathbf{x}_0$ and $\hat{\mathbf{x}}_0$. For adversarial training, the real sample $\mathbf{x}_{t^\prime}$ is obtained using the same forward diffusion process, while the fake sample $\hat{\mathbf{x}}_{t^\prime}$ is computed using $\mathbf{x}_t$ and the predicted velocity $\mathcal{G}_\theta\left(\mathbf{x}_t, t\right)$.}
  \label{fig: adversarial}

\end{figure}

%% file: sec/4_experiments.tex
\input{figs/4_qualitative}

\section{Experiments}
\label{sec:exp}

\input{tables/vbench}

\noindent\textbf{Training.}
The efficient image backbone is obtained by pruning the Stable Diffusion v1.5 UNet for $100K$ iterations on high-quality synthetic image datasets.
The model is then fine-tuned for $50K$ additional iterations to adapt to Rectified-Flow velocity prediction~\cite{sd3} as well as to the Spatial-Temporal VAE \cite{opensora}. 
We incorporate QK-norm and RoPE \cite{su2024roformer} in our network to stabilize training. 
The workflow for architecture search is discussed in \cref{sec: method search}. 
The image model pruning, temporal architecture search, and final model training are conducted on $256$ NVIDIA A100 80G GPUs using AdamW optimizer with $5e-5$ learning rate and betas values as $\left[0.9, 0.999\right]$.

\noindent\textbf{Adversarial Fine-tuning} is conducted for $6K$ iterations on $64$ NVIDIA A100 GPUs, using the AdamW optimizer with a learning rate of $1e-7$ for the generator (\ie, UNet) and $1e-4$ for the discriminator heads.
We set the betas as $\left[0.9, 0.999\right]$ for the generator optimizer, and $\left[0.5, 0.999\right]$ for the discriminator optimizer. We set the EMA rate as $0.95$ following SF-V~\cite{sfv}. We set $m=-1, s=1$ if not otherwise noted.

\noindent\textbf{Evaluation. }
The model is evaluated following the standard benchmarking procedure of VBench~\cite{huang2023vbench}. 
With the 4-step adversarial distilled model, we generate $120$-frame horizontal videos at a resolution of $432 \times 768$ using $4$ inference steps without employing classifier-free guidance. 
The generated video is saved at 5 seconds 24 FPS for score testing and qualitative visualization. 
The mobile demo and detailed demo settings are included in the \emph{supplementary material}. 

\subsection{Qualitative Visualization}
\label{sec:exp:qualitative}
We show visualizations of our generated videos in \cref{fig:qualitative}. 
Our model consistently produces high-quality video frames and smooth object movements. 
To demonstrate the generic text-to-video generation ability, we show various generation examples, including human, animal, photorealistic and art-styled scenes. 
We include more video visualizations in the \emph{supplementary material}. 

\subsection{Quantitative Comparisons}
\label{sec:exp:comp}
We present a comprehensive evaluation of our method against existing popular video generation models on VBench~\cite{huang2023vbench}, as in \cref{table: vbench comparison}.
Despite the fact that our model is compact and designated for fast inference on mobile, 
it achieves higher total score compared to recent arts, including the DiT-based OpenSora-V1.2, CogVideoX-2B~\cite{yang2024cogvideox}, and the UNet-based VideoCrafter-2.0 \cite{chen2024videocrafter2}.
In addition, compared to the 4-step distilled T2V-Turbo~\cite{li2024t2vturbo} and AnimateLCM ~\cite{wang2024animatelcm}, our model achieves better performance with more than $50\%$ reduction in size. 
The quantitative scores demonstrate the superiority of our efficient model design and the tailored adversarial distillation method. 

\noindent \textbf{User Study.}~
Human evaluations are conducted between our model and baselines as in the~\cref{table:user-study}.
We generate videos from VBench and Movie Gen Bench prompts and ask human labelers to pick the best results across \emph{prompt alignment}, \emph{aesthetics}, and \emph{motion}. 
The result shows that our model outperforms baseline metrics by a large margin.

\input{tables/user_study}

\subsection{Ablation Analysis}
\label{sec:exp:ablation}



\noindent \textbf{Comparison of Training Data Scheme.}
We compare the model trained with joint image-video datasets \vs video-only datasets.
As shown in \cref{table: ab training}, training with video-only datasets leads to significant performance degradation on the VBench score with a drop of 
$6.32$ in aesthetic quality,
$2.80$ in image quality,
and $2.54$ in total score.
The results highlight the importance of joint image-video training, as the image dataset offers more contextual information and enhances diversity.

\noindent \textbf{Model Scaling.}
We scale up and down the model size by adjusting the number of temporal layers, as shown in \cref{table: ab training}, to demonstrate the effectiveness of the proposed temporal architecture search. 
We can observe that scaling up the model can only marginally improve generative scores~(\ie, $\times2$ scale-up only increasing $0.56$ in dynamic degree and $0.12$ in motion smoothness, and $0.27$ in total score). 
However, both the $\times 2$ and $\times 4$ models hit the memory bound on iPhone. 
By dividing the $\times 2$ model into more chunks, we test its mobile speed and observe nearly doubled latency. 
While on the other hand, further scaling down the model results in heavy losses in generation quality~(\ie, decreasing $0.56$ in dynamic degree and $0.61$ in motion smoothness). 
Our efficient model is a balanced sweet point for quality and on-device performance. 

\input{tables/ablation_training}

\noindent \textbf{Effect of Different Temporal Layers.}
\input{figs/3_effect_temporal_modules}
To better understand the roles of the searched temporal layers, we compute the VBench score after systematically removing (i) all temporal layers in the downsample stage, (ii) bottleneck temporal layers, and (iii) all temporal layers in the upsample stage. 
The model makes reasonable zero-shot generations after temporal layer removal, but we still fine-tune 10K iterations under the same recipe as in \cref{sec: image video joint training} for fair comparison. 
As shown in \cref{fig: temporal modules}, removing different temporal layers results in varying degrees of performance degradation across different metrics, and all removing strategies result in a substantial drop in total score, demonstrating that the existence of the searched temporal layers is important, and they play different roles in generation. 
Specifically, removing temporal layers in upsample blocks results in a more significant loss in imaging quality, subject consistency, and background consistency, suggesting that the up temporal layers play important roles in detail reconstruction. 
In contrast, bottleneck layers are more important in human action and object class, where global information modeling dominates the results. 
We observe that removing down layers introduces less overall degradations compared to the other two, which is an anticipated phenomenon because the loss of modeling capacity can be mitigated by the subsequent bottleneck and up stage temporal layers after fine-tuning. 



\input{tables/ablation_adversarial}

\noindent \textbf{Effect of Discriminator Heads.}
We compare the effects of our spatial-temporal discriminator heads with the separate spatial and temporal heads proposed in SF-V~\cite{sfv} to demonstrate the effectiveness of our discriminator architecture.
For a fair comparison, both models are trained exclusively on video data. 
We evaluate the models fine-tuned with different discriminator heads on VBench~\cite{huang2023vbench}.
As shown in the first two rows of \cref{table: ab finetuning}, our discriminator heads result in improvements in both the quality score ($83.61$ \vs $83.60$) and the semantic score ($69.01$ \vs $64.25$) of the generated videos.

\noindent \textbf{Effect of Joint Image Video Fine-tuning.}
We examine the impact of incorporating image data during adversarial training, as shown in the second and fourth rows of \cref{table: ab finetuning}. 
The results indicate that fine-tuning the model with image data can slightly decrease the Quality score in the VBench~\cite{huang2023vbench} evaluation (from $83.61$ to $83.47$). 
However, by leveraging the increased diversity of the image dataset, the model achieves a substantial improvement in semantic performance, particularly in multi-object generation (from $37.85$ to $54.34$).
This enhancement leads to a better overall score compared to the model trained exclusively on video data ($81.14$ \vs $80.69$).

\noindent \textbf{Effect of Noise Distribution for Discriminator.}
Following \cref{equ:logitnorm}, the parameters $m$ and $s$ control the distribution of $t^\prime$, which determines the noise levels of $\mathbf{x}_{t^\prime}$ and $\hat{\mathbf{x}}_{t^\prime}$ before they are passed to the discriminator as real and fake samples, respectively. 
We investigate the effect of different noise distributions on model performance by evaluating the results using VBench~\cite{huang2023vbench}.
As shown in the last four rows of \cref{table: ab finetuning}, increasing $m$ (resulting in noisier real and fake samples) degrades the quality score (from $83.91$ to $77.84$) while slightly enhancing the temporal flickering score (from $99.29$ to $99.56$). 
Although setting $m=-2$ achieves the highest overall score among the experiments ($81.23$), it performs poorly on multi-object generation. 
Therefore, in most of our experiments, unless otherwise stated, we use $m=-1$. This setting yields a slightly lower overall score ($81.14$) but significantly improves semantic performance ($71.84$ \vs $70.54$) and excels in multi-object generation ($54.34$ \vs $47.64$).

\input{tables/ablation_steps}
\noindent \textbf{Effect of Inference Steps.}
Our fine-tuned model supports generation with a reduced number of inference steps. 
We further investigate how varying the number of evaluation steps affects the quality of the generated results, as shown in \cref{table: ab inference}. 
While our four-step generation achieves the best performance, even with only two steps, the model still produces reasonable results. 
Increasing the number of inference steps improves the performance of our model across all metrics. 
In \cref{table: ab inference}, we report not only the quality and semantic scores but also scores for dynamic degree, object class, and aesthetic quality.
However, reducing the process to a single inference step leads to a significant drop in performance. 
We leave more aggressive step reductions to future work.

%% file: figs/4_qualitative.tex
\begin{figure*}[ht]
  \centering
  \vspace{-.5em}
  \includegraphics[width=.9\linewidth]{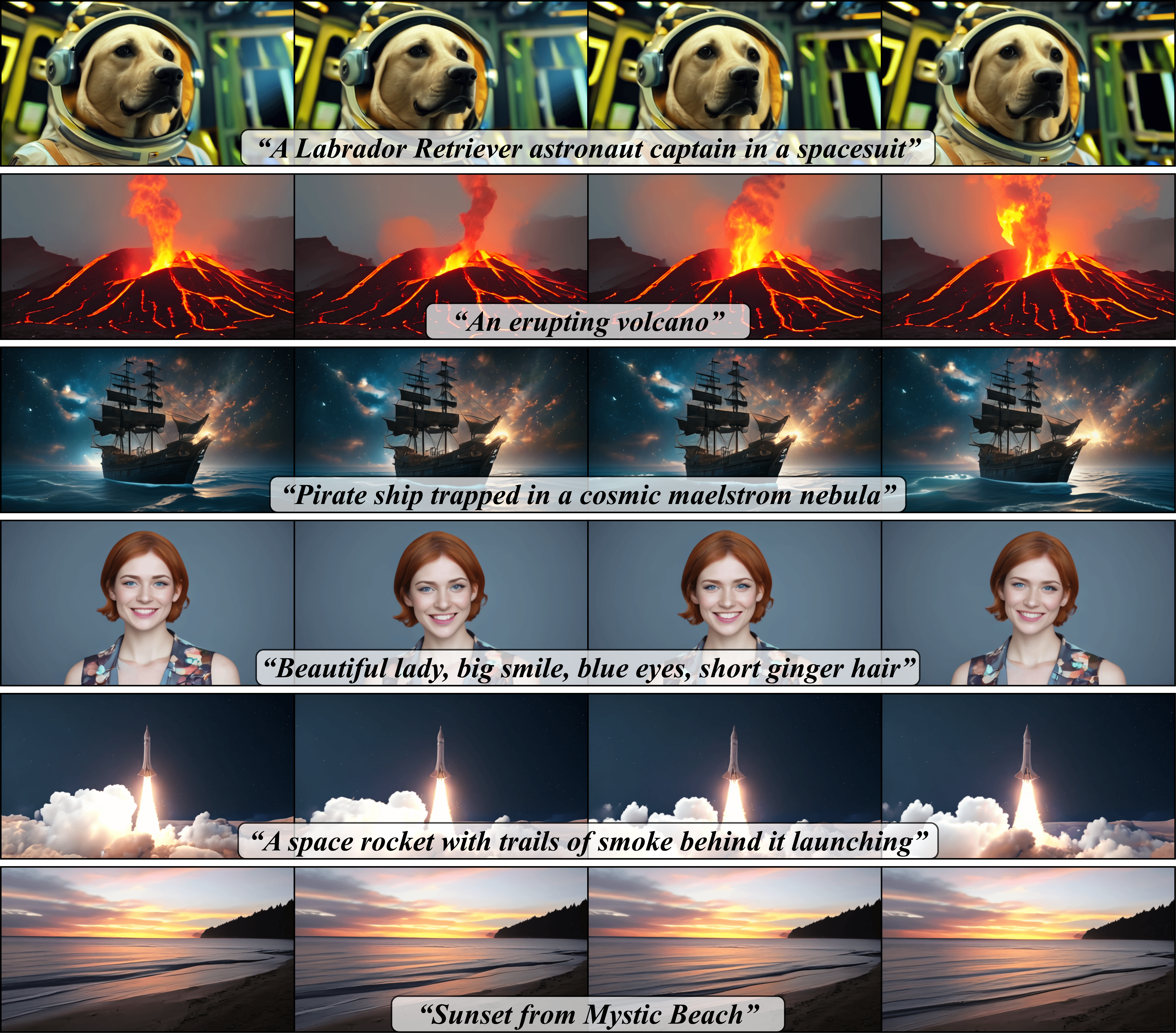}
  \caption{\textbf{Video generation on various domains.} We employ our model to synthesize videos across diverse domains, with each video containing $120$ frames at a resolution of $432\times 768$. All results are generated through a $4$-step inference process. The results demonstrate that our model can produce high-quality, motion-consistent videos featuring various objects across different domains.}
  \label{fig:qualitative}

  \vspace{-1.em}
\end{figure*}

%% file: tables/vbench.tex
\begin{table*}[htb!]
\small
\centering

\resizebox{0.95\linewidth}{!}{
\renewcommand\arraystretch{1}
\renewcommand\tabcolsep{1.25pt}
\begin{tabular}{l|c|c|c|ccccccccccccc}
\toprule
\multirow{2}{*}{Model} &
  Params &
  \multirow{2}{*}{Steps} &
  Total &
  Quality &
  Semantic &
  temporal &
  motion &
  dynamic &
  aesthetic &
  object &
  multiple &
  \multirow{2}{*}{color} &
  spatial &
  appearance &
  temporal &
  overall \\
                 & (B) &   & score & score & score & flickering & smoothness & degree & quality & class & objects &       & relationship & style & style & consistency \\
                 \hline
Kling            & --  &  -- & 81.85 & 83.39 & 75.68 & 99.30      & 99.40      & 46.94  & 61.21   & 87.24 & 68.05   & 89.90 & 73.03        & 19.62 & 24.17 & 26.42       \\
Pyramid Flow     &  2.0   & 20  & 81.72 & 84.74 & 69.62 & 99.49      & 99.12      & 64.63  & 63.26   & 86.67 & 50.71   & 82.87 & 59.53        & 20.91 & 23.09 & 26.23       \\
CogVideoX        & 5.0 & 50 & 81.61 & 82.75 & 77.04 & 98.66      & 96.92      & 70.97  & 61.98   & 85.23 & 62.11   & 82.81 & 66.35        & 24.91 & 25.38 & 27.59      \\
T2V-Turbo        &  1.6   &  4 & 81.01 & 82.57 & 74.76 & 97.48      & 97.34      & 49.17  & 63.04   & 93.96 & 54.65   & 89.90 & 38.67        & 24.42 & 25.51 & 28.16       \\
Emu3             &   8.0  &  --  & 80.96 & 84.09 & 68.43 & 98.57      & 98.93      & 79.27  & 59.64   & 86.17 & 44.64   & 88.34 & 68.73        & 20.92 & 23.26 & 24.79       \\
CogVideoX        & 1.6 & 50 & 80.91 & 82.18 & 75.83 & 98.89      & 97.73      & 59.86  & 60.82   & 83.37 & 62.63   & 79.41 & 69.90        & 24.80 & 24.36 & 26.66      \\
Pika-1.0         & --  &  -- & 80.69 & 82.92 & 71.77 & 99.74      & 99.50      & 47.50  & 62.04   & 88.72 & 43.08   & 90.57 & 61.03        & 22.26 & 24.22 & 25.94       \\
VideoCrafter-2.0 &  1.9 &  50 & 80.44 & 82.20 & 73.42 & 98.41      & 97.73      & 42.50  & 63.13   & 92.55 & 40.66   & 92.92 & 35.86        & 25.13 & 25.84 & 28.23       \\
AnimateDiff-V2   & 1.2 & 25 & 80.27 & 82.90 & 69.75 & 98.75      & 97.76      & 40.83  & 67.16   & 90.90 & 36.88   & 87.47 & 34.60        & 22.42 & 26.03 & 27.04      \\
OpenSora V1.2    & 1.2 & 30 & 79.76 & 81.35 & 73.39 & 99.53      & 98.50      & 42.39  & 56.85   & 82.22 & 51.83   & 90.08 & 68.56        & 23.95 & 24.54 & 26.85      \\
AnimateLCM       & 1.2 &  4 & 79.42 & 82.36 & 67.65 & 98.52      & 98.16      & 40.56  & 67.01   & 91.41 & 29.76   & 84.24 & 37.14        & 20.97 & 25.16 & 26.07       \\
ModelScope       &   1.4  &  50 & 75.75 & 78.05 & 66.54 & 98.28      & 95.79      & 66.39  & 52.06   & 82.25 & 38.98   & 81.72 & 33.68        & 23.39 & 25.37 & 25.67      \\ 
\hline
Ours           & 0.6 & 4 & 81.14 & 83.47 & 71.84 & 99.37      & 99.29      & 51.11  & 62.19   & 92.22 & 54.34   & 86.63 & 56.20        & 20.17 & 25.18 & 27.42      \\
\bottomrule
\end{tabular}
}
\caption{\textbf{Performance comparison with popular video generation models on VBench~\cite{huang2023vbench}.}}
\label{table: vbench comparison}
\vspace{-2em}
\end{table*}

%% file: tables/user_study.tex
\begin{table}[h]
\small
\centering
\scalebox{0.9}{
\begin{tabular}{c|ccc}
\toprule
Win Rate     & Prompt Alignment & Aesthetics & Motion \\
\hline
OpenSora-1.2 & 24.9\%             & 21.2\%       & 20.4\%   \\
CogVideoX-2B & 30.7\%             & 30.5\%       & 30.3\%   \\
Ours         & 44.4\%             & 48.3\%       & 49.3\%   \\
\bottomrule
\end{tabular}
}
\caption{\textbf{User Study} between OpenSora-1.2, CogVideoX-2B, and our model on \emph{prompt alignment}, \emph{aesthetics}, and \emph{motion}.}
\label{table:user-study}
\end{table}

%% file: tables/ablation_training.tex
\begin{table}[!htb]
\centering
\small
\resizebox{\linewidth}{!}{
\renewcommand\tabcolsep{2pt}
\begin{tabular}{l c c | c | c c c c|c}
\toprule

& \bf{w/ I}                   & \bf{Scaling} & $\Delta$\bf{T} & $\Delta$\bf{AQ} & $\Delta$\bf{IQ} & $\Delta$\bf{DD} & $\Delta$\bf{MS}
&  $\Delta$\bf{Total} \\
\midrule
Video-only & \ding{55}        & ---  & ---  & $-6.32$ & $-2.80$ & $+4.6$ & $-0.17$             
& $-2.54$ \\
\midrule
\multirow{3}{*}{Temporal Scaling}  & \multirow{3}{*}{\ding{51}}  
                              & $\times0.5$  & $-22\%$   & $-1.90$ & $+0.04$  & $-0.56$ & $-0.61$      
                              & $-1.58$ \\
&                             & $\times2$    & $+32\%$   & $+0.37$ & $+0.10$  & $+0.56$ & $+0.12$      
& $+0.27$ \\
&                             & $\times4$    & $+107\%$   & $+0.49$ & $-0.17$  & $+0.91$ & $+0.17$      
& $+0.31$ \\

\bottomrule
\end{tabular}
}
\caption{
\textbf{Analysis of training data scheme, latency, and quality of efficient architecture}.
The baseline model adopts the best suitable architecture and is trained with joint image-video datasets.
The ``scaling", and ``T" indicates the number of the temporal layers, and the latency comparing to the baseline.
The ``AQ", ``IQ", ``DD", and ``MS", are aesthetic quality, image quality, dynamic degree, and motion smoothness in VBench~\cite{huang2023vbench}.
The benchmark metrics are presented as the differences from the baseline model, where negative values indicates a decrease in performance over the baseline vice versa.
}
\label{table: ab training}

\end{table}

%% file: figs/3_effect_temporal_modules.tex
\begin{figure}
    \centering
    \vspace{-1.5em}
    \includegraphics[width=0.8\linewidth]{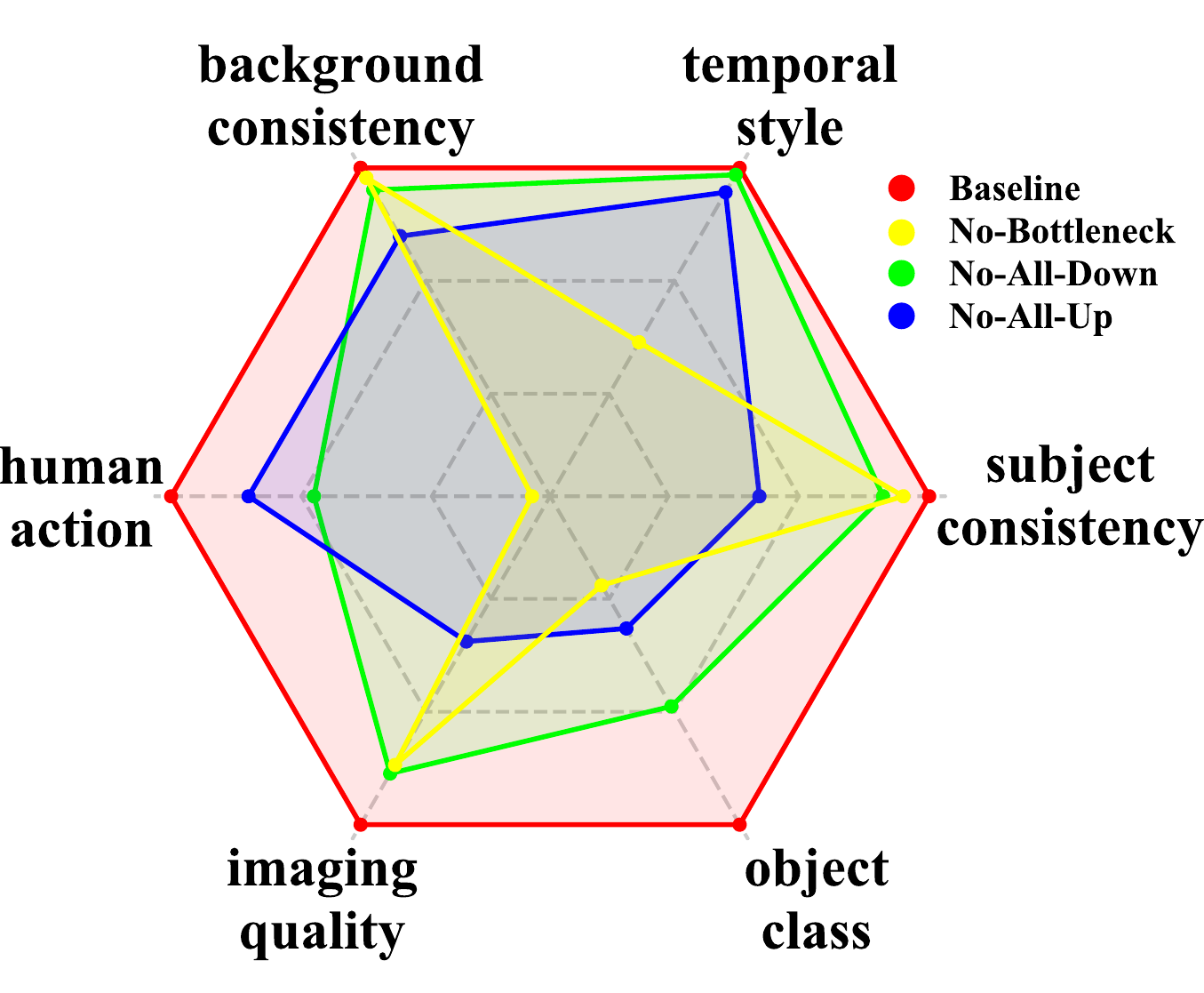}
    \caption{\textbf{Analysis of temporal layers.} We ablate temporal layers in different network stages to evaluate their effect. The scores are normalized according to our base model to better demonstrate the difference. 
    }
    \label{fig: temporal modules}
    \vspace{-1.em}
\end{figure}

%% file: tables/ablation_adversarial.tex
\begingroup
\setlength{\tabcolsep}{3.1pt} 
\renewcommand{\arraystretch}{0.9} 
\begin{table}[!htb]
\centering
\small
\resizebox{0.95\linewidth}{!}{
\begin{tabular}{l c c r | c c c c | c}
\toprule

\bf{Head} & \bf{w/ I} & \bf{w/ $\mathcal{L}_{\text{adv}}$} & $m$&  \bf{TF} & \bf{MO} &   \bf{Q} & \bf{S} &  \bf{Total} \\
\midrule
SF-V~\cite{sfv} & \multirow{2}{*}{\ding{55}} & \multirow{2}{*}{\ding{51}}& \multirow{2}{*}{$-1$}& $98.86$& $30.75$&  $83.60$ & $64.25$ & $79.73$ \\
Ours & & & & $99.27$& $37.85$ &  $\underline{83.61}$ & $69.01$ & $80.69$ \\

\midrule
Ours & \ding{51} & \ding{55} &$-1$& $97.32$& $1.01$ &   $70.03$ & $17.04$ & $59.44$ \\

\midrule
\multirow{4}{*}{Ours} &\multirow{4}{*}{\ding{51}} &\multirow{4}{*}{\ding{51}} & $-2$ & $99.29$& $\underline{47.64}$ &   $\bf{83.91}$ & $\underline{70.54}$ & $\bf{81.23}$ \\
 & & &  \cellcolor{LightCyan}$-1$ & \cellcolor{LightCyan}$99.37$& \cellcolor{LightCyan}$\bf{54.34}$ &  \cellcolor{LightCyan}$83.47$ & \cellcolor{LightCyan}$\bf{71.84}$  & \cellcolor{LightCyan}$\underline{81.14}$ \\
& & &$0$& $\underline{99.44}$& $43.63$ &   $81.69$ & $69.46$ & $79.24$ \\
& & &$1$ & $\bf{99.56}$& $24.41$&   $77.84$ & $58.88$ & $74.05$ \\
    \bottomrule
\end{tabular}
}
\caption{\textbf{Analysis of our adversarial fine-tuning scheme.} We evaluate the VBench~\cite{huang2023vbench} scores for our models using different training schemes. In the results, ``TF'' and ``MO'' denote the temporal flickering and multiple objects sub-scores, respectively, while ``Q'' and ``S'' represent the quality and semantic scores. The table summarizes how varying (1) the type of discriminator head, (2) training with or without an image dataset, and (3) different $t^\prime$ distributions impact the performance of our model.}
\label{table: ab finetuning}

\end{table}
\endgroup

%% file: tables/ablation_steps.tex

\begingroup
\setlength{\tabcolsep}{3.1pt} 
\renewcommand{\arraystretch}{0.9} 
\begin{table}[!htb]
\centering
\small
\begin{tabular}{l | c c c c c c}
\toprule

\bf{\# Steps}& \bf{DD}& \bf{OC} & \bf{AQ} & \bf{Quality} & \bf{Semantic} &  \bf{Total} \\
\midrule
$1$ & $11.39$ & $79.24$& $52.85$ & $74.43$ & $62.70$ & $72.09$ \\
$2$ & $30.28$ & $88.97$& $59.74$ & $80.08$ & $69.28$ & $77.92$ \\
$4$ & $\bf{51.11}$ & $\bf{92.22}$& $\bf{62.19}$ & $\bf{83.47}$ & $\bf{71.84}$  & $\bf{81.14}$ \\
    \bottomrule
\end{tabular}
\caption{\textbf{Analysis of the number of inference steps.} We measure VBench~\cite{huang2023vbench} score with different numbers of inference steps. In the results, ``DD", ``OC", and ``AQ" denote the dynamic degree, object class, and aesthetic quality scores, respectively.}
\label{table: ab inference}
\end{table}
\endgroup




%% file: sec/5_conclusion.tex
\section{Discussion and Conclusion}
\label{sec:conclusion}

In this work, we propose an acceleration framework for the video diffusion model, and for the first time, achieve super-fast text-to-video generation on mobile devices. 
Specifically, we discover an efficient but powerful network architecture through latency and memory joint architecture search for temporal layers. In addition, we propose an improved adversarial fine-tuning technique to distill our model to $4$ steps to further speed up generation. 
Our work is a good starting point for the edge deployment of video diffusion models and we hope to inspire more downstream applications such as video extension and editing. 

\noindent \textbf{Limitations.} We use a public $4$-channel VAE ~\cite{opensora} to encode videos to latent space. Recent work has shown that using more latent channels benefits reconstruction details. 
Another future direction is to further improve the step reduction technique for 1-2 denoising steps, as works ~\cite{sfv} on server-level models. 

%% file: sec/X_suppl.tex
\clearpage
\appendix
\setcounter{page}{1}
\maketitlesupplementary

\setcounter{section}{0}
\setcounter{figure}{0}
\setcounter{table}{0}
\renewcommand{\thefigure}{A\arabic{figure}}
\renewcommand{\thetable}{A\arabic{table}}
\renewcommand{\thealgorithm}{A\arabic{algorithm}}

\section*{Overview}

The supplementary material accompanying this paper provides additional insights and elaborations on various aspects of our proposed method.
The contents are organized as follows:

\begin{itemize}
    \item \textbf{Search Algorithm:} \cref{sec: supp: search} provides a detailed search algorithm we proposed to determine the final architecture of our model.
    \item \textbf{VAE Compression:} \cref{sec: supp: vae} describes the separable efficient variational autoencoder we employed for efficient video generation on mobile devices.
    
    \item \textbf{Qualitative and Quantitative Results for the Spatial Backbone:} We showcase a broad range of qualitative results demonstrating the effectiveness of our spatial backbone. The quantitative results is also evaluated. The results can be found in \cref{sec: supp: qualitative spatial}.
    
    \item \textbf{Qualitative Comparison Results:}
     The qualitative comparison of our model with two popular open-source models~(OponSora-v1.2~\cite{opensora} and CogVideoX-2B~\cite{yang2024cogvideox}) is shown in \cref{sec: supp: comp}.
     More results can be found in the \emph{accompanying webpage}.
     \item \textbf{More Qualitative Results:} Additional qualitative results are presented in \cref{sec: supp: qualitative}. We also provide these results in video format in the \emph{accompanying webpage}.
    
    \item \textbf{Demo:} We provide our demo benchmark and mobile screenshots in \cref{sec: supp: demo}.

    \item \textbf{Effect of Adversarial Fine-tuning}: We further discuss the effect of adversarial finetuning for step distillation in~\cref{sec: supp: adv-ft}.

    \item \textbf{Latency Analysis}: \cref{sec: supp: latency} shows the latency analysis of different temporal blocks.
    
\end{itemize}

\section{Search Algorithm}
\label{sec: supp: search}
We propose a two-step architecture search to design temporal layers that satisfy hardware constraints and performance requirements.
First, a coarse architecture search is conducted based on the spatial backbone, eliminating candidate architectures that violate the hardware constraints to narrow the search space.
Then, we build an action set, $\mathcal{A}\in \{A^{+,-}_{\textit{SelfAttnND}[i]}, A^{+,-}_{\textit{CrossAttnND}[i]}, A^{+,-}_{\textit{ConvND}[i]}\}$, to perform the evolutionary search, where the $A^{+,-}$ indicates the action to add or remove the temporal layer for corresponding position~($i^{th}$ block).
The action is guided by latency and memory constraints, as well as generation performance.
We choose the Vbench score~\cite{huang2023vbench} to evaluate the quantitative performance of each architecture, and we specifically focus on the average score of the \textit{overall consistency}, the \textit{object class}, and the \textit{color} score instead of the complete benchmark to reduce the evaluation time.
The value score of each action is defined as $\{\frac{\Delta \text{Vbench}}{\Delta \text{Latency}},\frac{\Delta \text{Vbench}}{\Delta \text{Memory}}\}$.
We use $268$ prompts with $25$ denoising steps and $7$ classifier-free guidance scale to benchmark those scores above in Vbench~\cite{huang2023vbench}, and it takes $8$ A100 GPU hours to evaluate each action.
We further simplify the search space by avoiding a mixture of temporal layers in the same position.
As shown in \cref{algorithm: search}, different temporal layers are integrated into the UNet at each search step, with evaluations based on the selected Vbench score after training the model for $20K$ iterations.
The latency and peak memory are retrieved from the pre-built look-up table.
The action is then updated based on the $\frac{\Delta \text{Vbench}}{\Delta \text{Latency}}$ and $\frac{\Delta \text{Vbench}}{\Delta \text{Memory}}$, prioritizing temporal layers that offer low latency and memory consumption while contributing more significantly to a better Vbench score.
\input{algo/algo_search}

\input{figs/rebuttal_prune_unet}
\section{VAE Compression}
\label{sec: supp: vae}
\noindent \textbf{Separable Variational Autoencoder}.
The variational auto-encoder~(VAE) decoder for video is more time-consuming and memory-intensive than its image counterpart, as it processes a sequence of frames as inputs.
To mitigate memory consumption, we disentangle the spatial and temporal decoders to mitigate memory consumption.
Specifically, given a latent feature $\mathbf{x}_0 \in \mathbb{R}^{\tilde{n} \times 4 \times \tilde{H} \times \tilde{W}}$, the $\mathbf{x}_0$ is first decoded to $\mathbf{x}_{t0} \in \mathbb{R}^{n \times 4 \times \tilde{H} \times \tilde{W}}$ by the temporal decoder, and then decoded back to pixel space $\mathbf{v} \in \mathbb{R}^{n \times 3 \times H \times W}$ by the spatial decoder.
This approach allows us to split the latent feature $\mathbf{x}_0$ into multiple sub-features for inference, significantly reducing the peak memory.
For example, a latent feature $\mathbf{x}_0 \in \mathbb{R}^{\tilde{n} \times 4 \times \tilde{H} \times \tilde{W}}$ can be sliced to multiple sub-features with dimension $\tilde{n}^\prime \times 4 \times \tilde{H} \times \tilde{W}$, where $\tilde{n}^\prime < \tilde{n}$, then fed into the temporal decoder.
Similarly, the temporal reconstructed latent feature, with dimension $n \times 4 \times \tilde{H} \times \tilde{W}$, can also be fed into the spatial decoder with smaller segments such as $1 \times 4 \times \tilde{H} \times \tilde{W}$.
This approach balances memory consumption, memory I/O, and GPU/NPU utilization, promising hardware-friendly inference.

\noindent \textbf{VAE Decoder Compression}.
We conduct VAE compression only on the decoder to speed up the inference process.
The encoder weights are frozen during the compression, and we only train the decoder.
We replace the convolution in the original decoder with depth-wise separable convolution for better I/O and less computation.
Moreover, a distill loss is adopted to maintain the reconstruction quality of the decoder.
The quality comparison is shown in \cref{table: vae decoder}, which demonstrates our efficient decoder can achieve $\times 54.5$ speed-up with even better performance.

\begin{table}[ht]
\small
\centering
\resizebox{\linewidth}{!}{
\begin{tabular}{c|ccccc}
\toprule
VAE      & Latency (s) & PSNR  & SSIM   & LPIPS  & FloLPIPS \\
\hline
OpenSora & 27.2      & 29.07 & 0.8066 & 0.1336 & 0.1303   \\
Ours     & 0.5       & 29.21 & 0.8240 & 0.0949 & 0.0915   \\
\bottomrule
\end{tabular}
}
\caption{\textbf{Our VAE with efficient decoder.} }
\label{table: vae decoder}

\end{table}

\section{Qualitative and Quantitative Results for the Spatial Backbone}
\label{sec: supp: qualitative spatial}

We present the qualitative results of our efficient spatial backbone, as shown in \cref{fig:supp:image-quality}.
These images demonstrate that our spatial backbone can achieve high-fidelity text-to-image generation quality, which promises text-to-video generation quality.
We compare the CLIP-score and aesthetic score of our model with the Stable Diffusion v1.5~\cite{rombach2022high}. The evaluation is conducted on a subset of $6000$ images from the MS-COCO 2014 validation set.
As shown in \cref{table: spatial backbone}, our model achieves $\times2.5$ compression rate while delivering better CLIP-score~($0.33$ \vs $0.31$) and aesthetic score~($6.23$ \vs $5.51$), exhibiting its impressive text-to-image generation quality.

Additionally, we exhibit the quality comparison of our spatial backbone with SD1.5 in~\cref{fig:rebuttal:prune-unet}.

\begin{table}[ht]
\small
\centering
\resizebox{\linewidth}{!}{
\begin{tabular}{c|ccc}
\toprule
Model    & Params (M)  & CLIP-Score~$\uparrow$ & Aesthetic Score~$\uparrow$  \\
\hline
SD v1.5  & 820 & 0.31 & 5.51  \\
Ours     & 327 & 0.33 & 6.23  \\
\bottomrule
\end{tabular}
}
\caption{\textbf{Quantitative Results of Our Spatial backbone.} }
\label{table: spatial backbone}

\end{table}

\section{Qualitative Comparison Results}
\label{sec: supp: comp}
The comparison of our model with OpenSora-v1.3\cite{opensora} and CogVideoX-2B~\cite{yang2024cogvideox} is shown in \cref{fig:supp:comparison}.
More comparisons are presented in \href{https://snap-research.github.io/snapgen-v/}{project page}.

\section{More Qualitative Results}
\label{sec: supp: qualitative}

In this section, we present an extensive collection of qualitative results, as shown in \cref{fig:supp:qualitative}, that demonstrate the capabilities of our proposed method.
This includes both the examples showcased in the main paper and additional results, offering a comprehensive view of our method's performance in various scenarios. 

To facilitate a more interactive and illustrative experience, these qualitative results are provided in video format.
Readers are recommended to check these results in \href{https://snap-research.github.io/snapgen-v/}{project page}.
This visualization provides a more nuanced understanding of the temporal and visual qualities of our method.

\section{Demo Settings}
\label{sec: supp: demo}

Our demo is evaluated on an iPhone 16 Pro Max, equipped with an Apple A18 Pro chipset featuring a six-core CPU, six-core GPU, and 16-core Neural Engine.
Our model is converted to FP16 and executed on the Neural Engine and the CPU cores.
To enhance efficiency, timestep embeddings are also pre-computed since these values are fixed for each timestep. 
The inference pipeline takes four denoising steps without classifier-free guidance. 
To enable a fast mobile demo with pleasing quality, we adjust the input size for denoiser to $15\times64\times64$, which yields an output video clip with $51$ frames $512\times512$ resolution.
To ensure the video quality, the model is further finetuned with video datasets with a framerate of $10$~fps.
Hence, the $51$ frame clip is $5.1$ seconds in length.
Our UNet model is exported by CoreML and benchmarked using Xcode Performance tools.
Furthermore, the exported model is split into two parts for loading and execution efficiency.
The latency benchmark screenshots are shown in~\cref{fig:supp:demo_xcode}, thus one denoising step takes 1.02 seconds.
Similarly, the text-encoder and VAE-decoder take 6 ms and 0.5 seconds, respectively.
Thus, the entire inference pipeline takes less than 5 seconds on average.
We exhibit the mobile demo screenshots in~\cref{fig:supp:demo_screenshot} and \href{https://snap-research.github.io/snapgen-v/}{project page}. 

\section{Effect of Adversarial Fine-tuning}
\label{sec: supp: adv-ft}
\cref{table: ab finetuning} also shows the effect of adversarial fine-tuning. Tuning without adversarial loss can not yield promising results compared to the baseline for step distillation.

\section{Latency Analysis}
\label{sec: supp: latency}
\input{figs/supp_latency}
The latency of different temporal blocks is shown in \cref{fig:supp: latency}.


\input{figs/supp_image_quality}

\input{figs/supp_comparison}

\input{figs/supp_qualitative}
\input{figs/supp_demo}



%% file: algo/algo_search.tex
\begin{algorithm}[ht]
\small
\caption{Search Algorithm}
\label{algorithm: search}
\begin{algorithmic}
\Require  
{\\ UNet: $\hat{{\epsilon}}_{\theta}$;\\
validation set: $\mathbb{D}_{\text{val}}$; \\
latency and memory lookup table  $\mathbb{T}:$  \\
\quad $\{\textit{SelfAttnND}[i]$, $\textit{CrossAttnND}[i]$, $\textit{ConvND}[i]\}$  .}
\Ensure $\hat{{\epsilon}}_{\theta}$ converges and satisfies latency objective $S$. 

\While{$\hat{{\epsilon}}_{\theta}$ not converged}
\State $\rightarrow$ \textbf{Architecture optimization}:
\If{perform architecture evolving at this iteration}
\State $\rightarrow$ \textbf{Evaluate blocks}:
\For{each $\text{block}[i]$}
\State {\small $\Delta \text{Vbench} \leftarrow \operatorname{eval}(\hat{{\epsilon}}_{\theta}, A^-_{\text{block}[i]}, \mathbb{D}_{\text{val}})$}, 
\State {\small $\Delta \text{Latency}, \Delta \text{Memory} \leftarrow \operatorname{eval}( \hat{{\epsilon}}_{\theta}, A^-_{\text{block}[i]}, \mathbb{T})$}
\EndFor
\State $\rightarrow$ \textbf{Sort actions based on} $\frac{\Delta \text{Vbench}}{\Delta \text{Latency}}$ \textbf{and} $\frac{\Delta \text{Vbench}}{\Delta \text{Memory}}$ \textbf{, execute action, and evolve architecture to get latency} $T$ \textbf{and peak memory} $M$: 
\If{$T$ not satisfied}
\State {\small $\{\hat{A}^-\} \gets {\arg\min}_{A^-} \frac{\Delta \text{Vbench}}{\Delta \text{Latency}}$,}
\ElsIf{$M$ not satisfied}
\State {\small $\{\hat{A}^-\} \gets {\arg\min}_{A^-} \frac{\Delta \text{Vbench}}{\Delta \text{Memory}}$,}
\Else
\State {\small $\{\hat{A}^+\} \gets \operatorname{add}({\arg\max}_{A^-} \{\frac{\Delta \text{Vbench}}{\Delta \text{Latency}},\frac{\Delta \text{Vbench}}{\Delta \text{Memory}}\})$,}
\State {\small $\hat{{\epsilon}}_{\theta} \gets \operatorname{evolve}( \hat{{\epsilon}}_{\theta}, \{\hat{A}\} )$}
\EndIf
\EndIf
\EndWhile
\end{algorithmic}   
\end{algorithm}

%% file: figs/rebuttal_prune_unet.tex
\begin{figure}[ht]
  \centering
  \includegraphics[width=\linewidth]{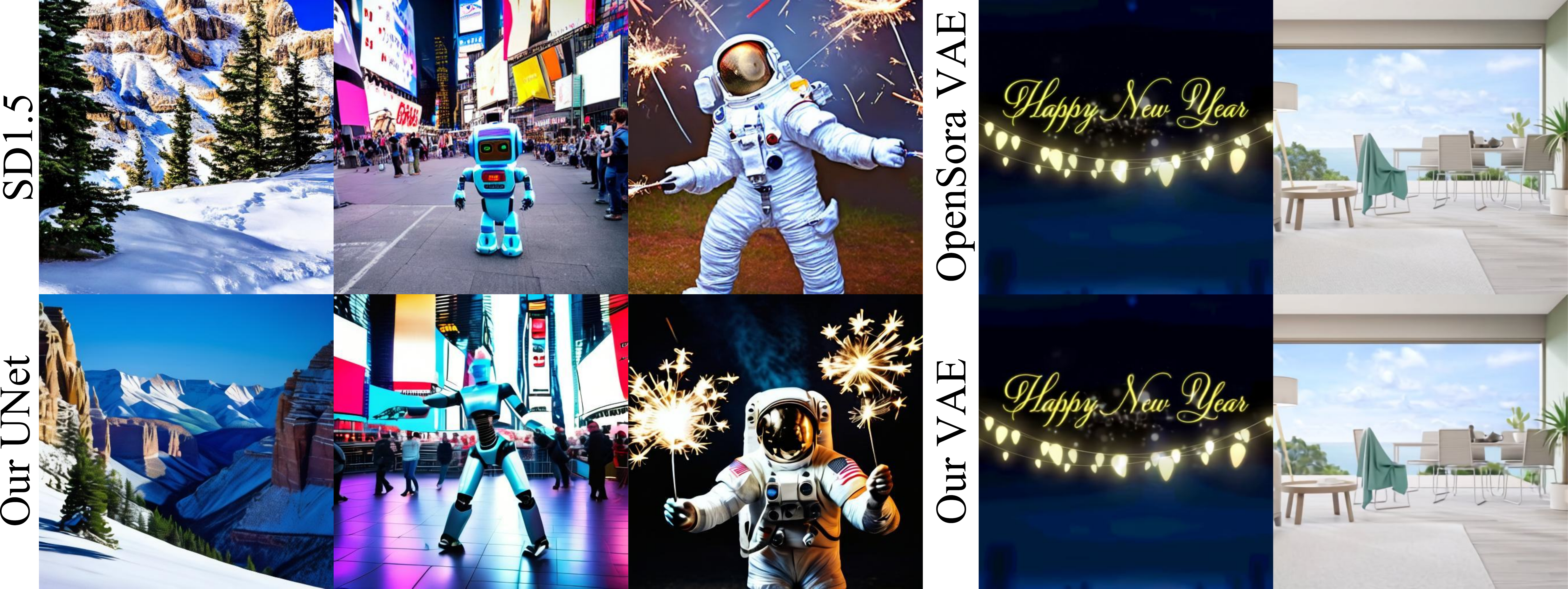}
  \caption{Comparison between the SD1.5 and our efficient spatial UNet backbone.}
  \label{fig:rebuttal:prune-unet}
\end{figure}

%% file: figs/supp_latency.tex
\begin{figure}
    \centering
    \includegraphics[width=0.75\linewidth]{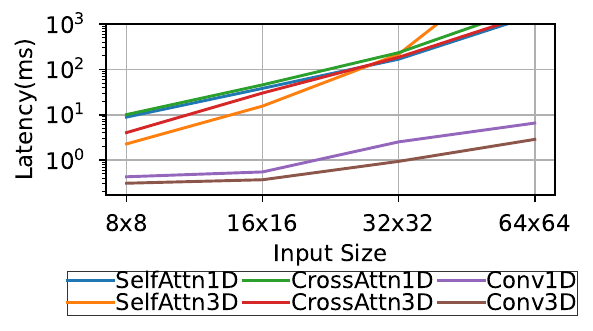}
    \caption{Latency}
    \label{fig:supp: latency}
\end{figure}

%% file: figs/supp_image_quality.tex
\begin{figure*}[ht]
    \centering
    \captionsetup{type=figure}
    \includegraphics[width=0.95\linewidth]{assets/supp_image_quality.pdf}
    \caption{\textbf{Qualitative results of the spatial backbone.}}
    \label{fig:supp:image-quality}
\end{figure*}

%% file: figs/supp_comparison.tex
\begin{figure*}[ht]
  \centering
    \captionsetup{type=figure}
  \includegraphics[width=0.9\linewidth]{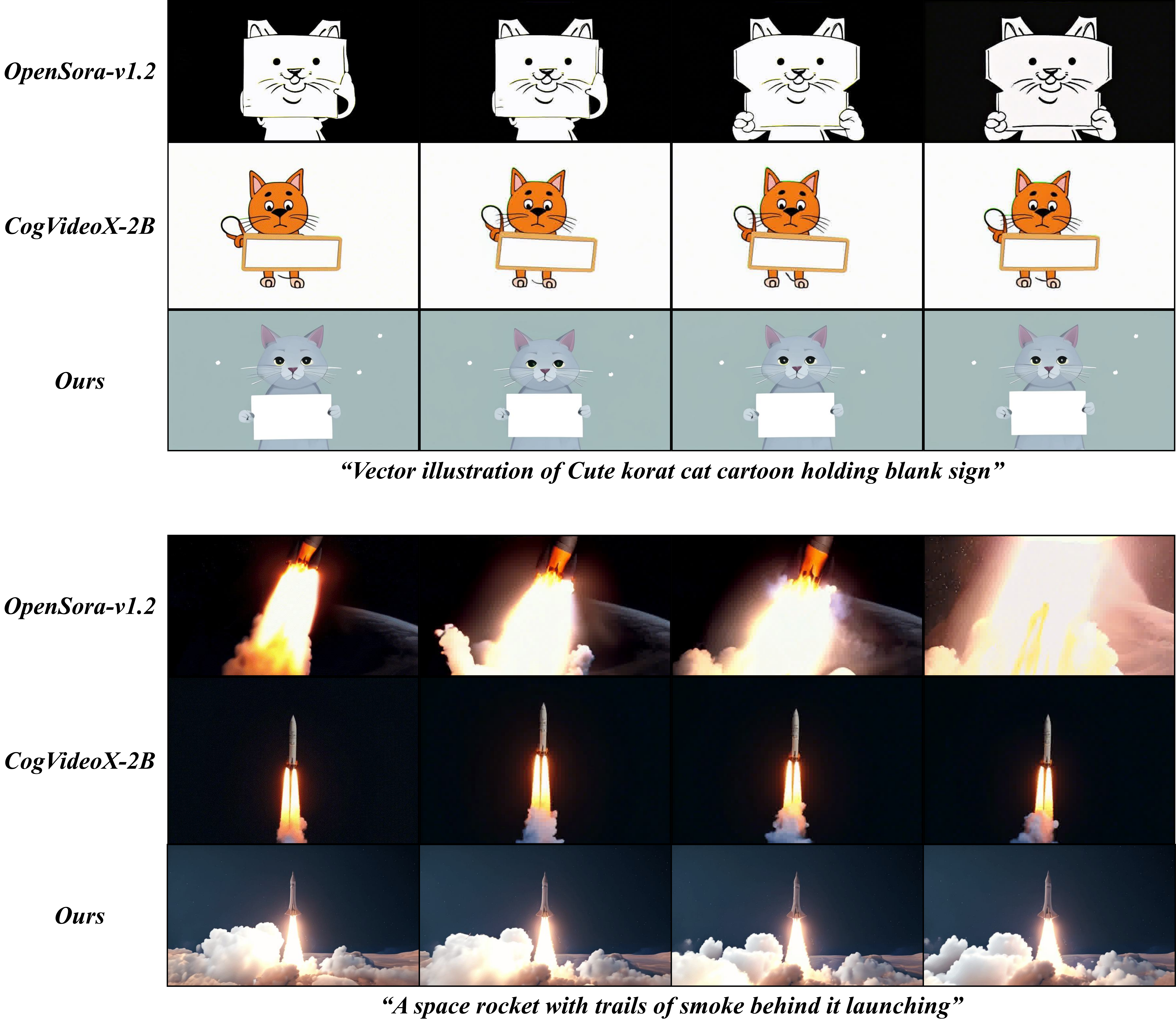}
  \caption{\textbf{Comparison with OpenSora-v1.2~\cite{opensora} and CogVideoX-2B~\cite{yang2024cogvideox}.}}
  \label{fig:supp:comparison}
\end{figure*}

%% file: figs/supp_qualitative.tex
\begin{figure*}[ht]
  \centering
  \captionsetup{type=figure}
  \includegraphics[width=0.65\linewidth]{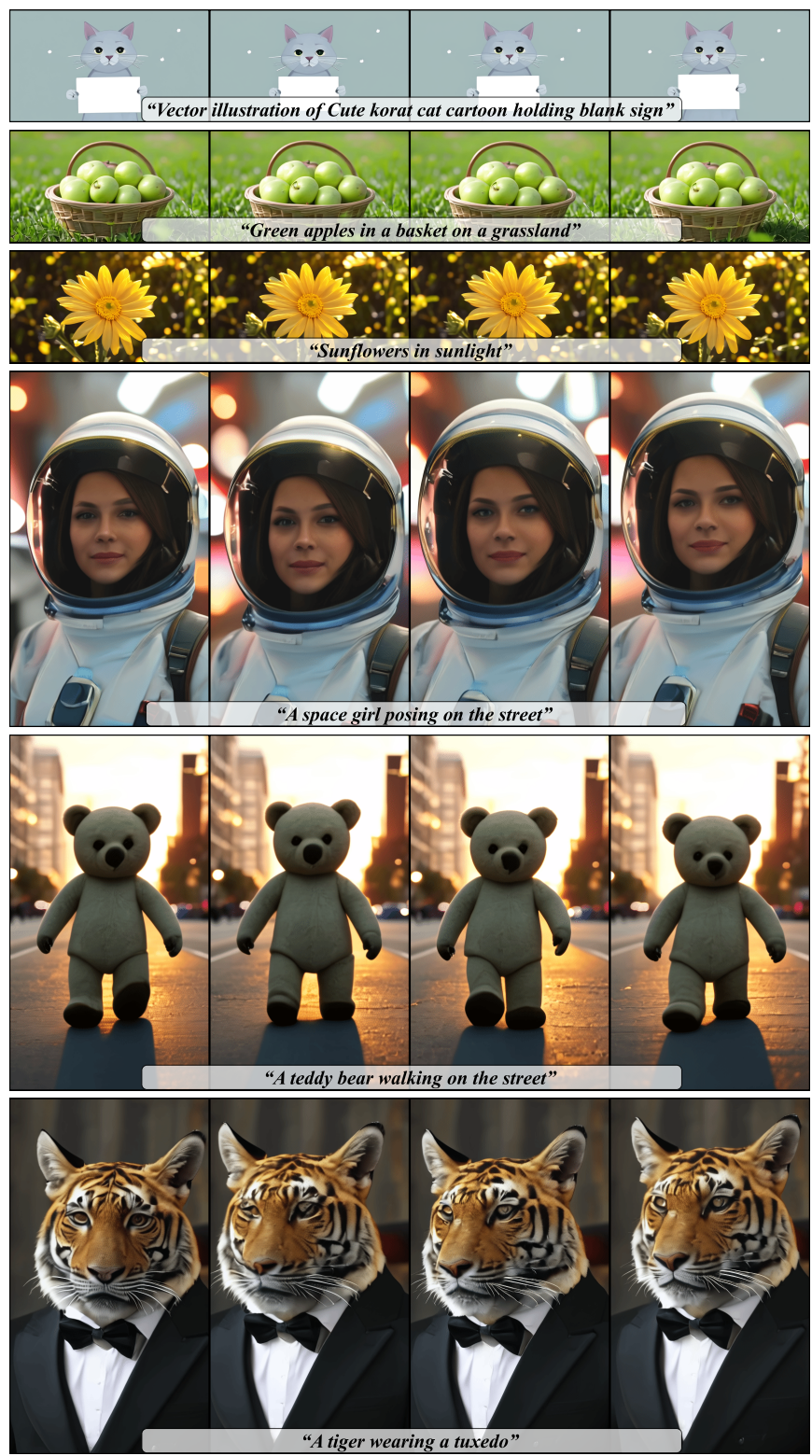}
  \caption{\textbf{More qualitative results.}}
  \label{fig:supp:qualitative}
\end{figure*}

%% file: figs/supp_demo.tex
\begin{figure*}[ht]
  \centering
  \captionsetup{type=figure}
    \includegraphics[width=0.95\linewidth]{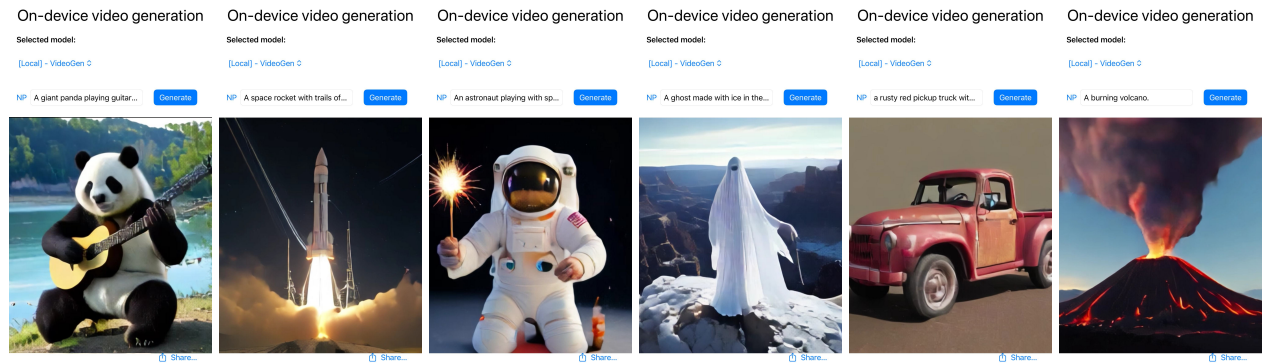}
  \caption{\textbf{Screenshots of Mobile Demo.}}
  \label{fig:supp:demo_screenshot}
\end{figure*}

\begin{figure*}[ht]
  \centering
  \captionsetup{type=figure}
    \includegraphics[width=0.95\linewidth]{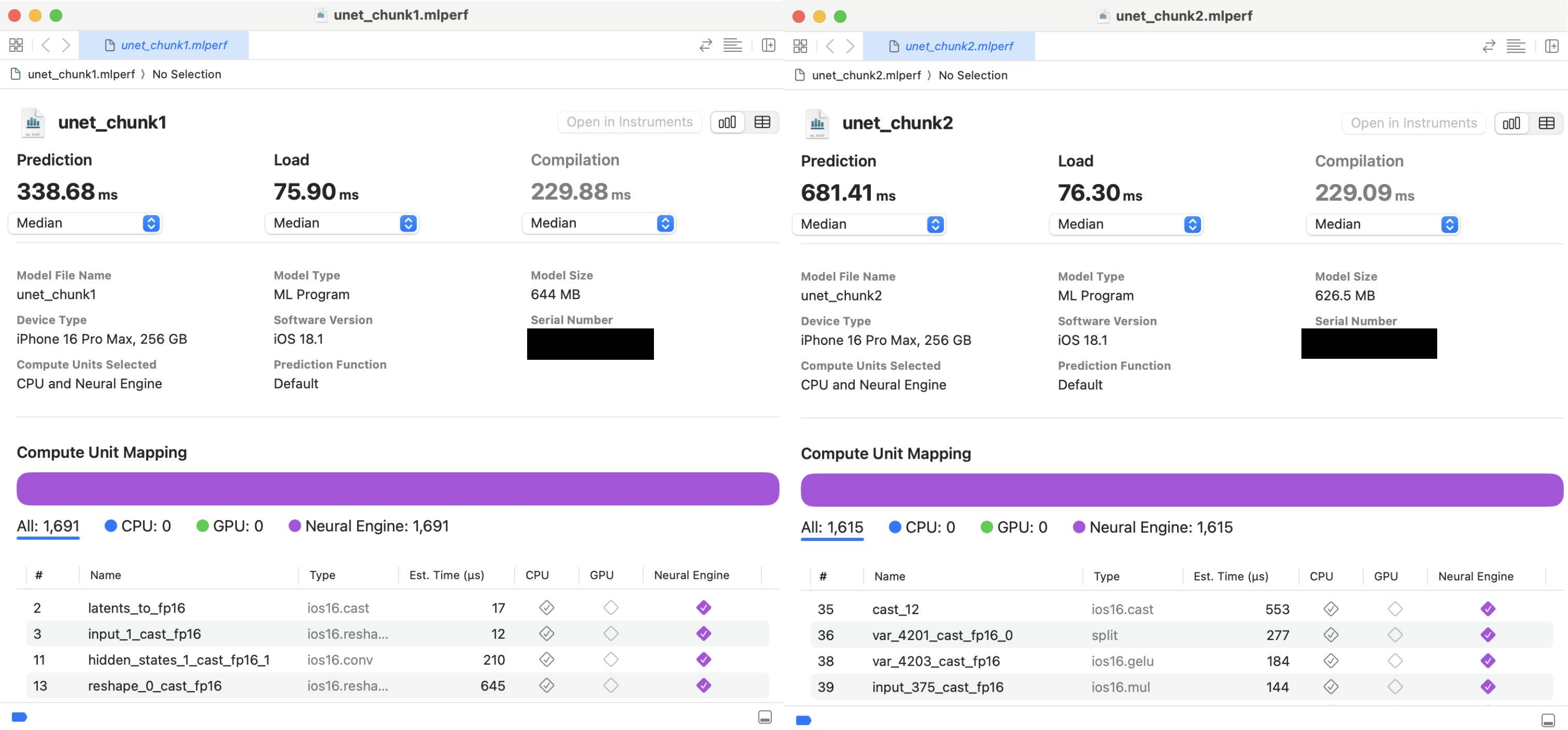}
  \caption{\textbf{UNet Latency Benchmark on iPhone 16 Pro Max.}}
  \label{fig:supp:demo_xcode}
\end{figure*}